\documentclass{article} 

\usepackage{iclr2027_conference,times}

\usepackage{amsmath,amsfonts,bm}

\def\eqref#1{equation~\ref{#1}}

\def\1{\bm{1}}

\DeclareMathAlphabet{\mathsfit}{\encodingdefault}{\sfdefault}{m}{sl}
\SetMathAlphabet{\mathsfit}{bold}{\encodingdefault}{\sfdefault}{bx}{n}

\usepackage{microtype}
\usepackage{graphicx}
\usepackage{subcaption}
\usepackage{hyperref}
\hypersetup{hidelinks}
\usepackage{multirow}
\usepackage{mathtools}
\usepackage{tablefootnote}
\usepackage{threeparttable}

\usepackage{url}
\usepackage{booktabs}       

\usepackage{algorithm}
\usepackage[noend]{algpseudocode}
\usepackage{makecell}

\usepackage{wrapfig}

\title{SlimWise: Decoupling Expert Pruning Across Prefill and Decode for Efficient MoE Serving}

\author{Gunho Park$^1$, Kyoungho Jeun$^1$, Juntaek Oh$^1$, Byeongjun Shin$^{1,2}$, Baeseong Park$^1$,  \\
 \textbf{Minsoo Rhu}$^{1,2}$ \\
$^1$ a2sys, $^2$ KAIST\\
\texttt{\{gunho.park, minsoo.rhu\}@a2sys.ai}
}

\newcommand{\dk}[1]{{\scriptsize\color{gray}(#1)}}

\iclrfinalcopy 
\begin{document}


\maketitle

\begin{abstract}
{
Mixture-of-experts (MoE) models activate few experts per token, yet batched decoding can access nearly the entire expert pool, making expert-weight traffic a major bottleneck. 
Expert pruning reduces this traffic, but conventional approaches also prune compute-bound prefill, sacrificing model quality for little throughput benefit. 
We present SlimWise, a serving framework that tailors the expert pool to each inference phase. SlimWise performs prefill with the full model and decode with a pruned model that directly reuses the prefill-generated KV cache without conversion.
Across two MoE backbones and three pruning criteria, this training-free KV cache handoff substantially narrows accuracy gaps relative to the full model in many settings. 
We also show that benchmark accuracy can conceal substantial pruning-induced changes in generation length. 
To address these distortions and residual accuracy loss, SlimWise introduces a low-cost distillation stage that trains the decoder to continue from full-model KV caches while updating only a small subset of parameters. 
Implemented in vLLM, SlimWise supports both prefill–decode (PD) disaggregation and PD-colocated serving. 
On Qwen3.6-35B-A3B, SlimWise improves decode throughput by up to 1.81$\times$ at 50\% expert pruning with minimal accuracy loss.
}
\end{abstract}

\section{Introduction}
\label{sec:intro}

Sparse mixture-of-experts (MoE) models increase model capacity while limiting per-token computation by activating only a small fraction of a large expert pool ~\citep{fedus2022switch, jiang2024mixtral, liu2024deepseek, yang2025qwen3}. This per-token computational sparsity, however, does not guarantee low memory traffic under continuous batching, which processes multiple sequences per inference step~\citep{yu2022orca, kwon2023efficient}. Each step loads the union of experts selected across the batch, which can approach the full expert pool as batch size grows~\citep{gupta2024lynx, oncescu2025opportunistic, vankov2026xshare}. 
This expert-weight traffic has different performance implications for prefill and decode. 
Prefill processes many prompt tokens simultaneously, allowing each expert’s weights to be reused across enough tokens to make execution compute-bound. Consequently, reducing the expert pool while keeping the number of active experts per token unchanged offers little performance benefit. Decode, by contrast, produces only one token per sequence in each step, leaving too few tokens per expert to amortize the cost of loading its weights. Increasing the batch size helps, but KV cache capacity and per-token latency requirements limit the concurrency that serving systems can sustain~\citep{zhang2025janus, agrawal2024taming}. Expert-weight movement thus remains a major bottleneck in batched MoE decoding.

Two complementary approaches address this bottleneck.
One reduces per-batch expert access through token rerouting, expert folding, or locality-aware request routing while retaining the full expert pool~\citep{gupta2024lynx, oncescu2025opportunistic, wu2026exfold, choi2026eldr}. The other reduces the pool itself through pruning or merging ~\citep{lasby2026reap, jaiswal2025finding, chen2025hcsmoe, xie2024moe}. 
Expert pruning ranks experts using importance statistics from calibration rollouts and removes those with the lowest scores. These approaches target complementary sources of cost: the former reduces the fraction of the expert pool activated in each inference step, while the latter reduces the total number of experts retained during inference.

In this work, we focus on expert pruning, which is effective in reducing expert-weight traffic but has two limitations.
First, the effect of pruning on model quality degradation varies across pruning criteria and benchmarks~\citep{lasby2026reap}. The impact of pruning also extends beyond accuracy: a pruned model can retain high benchmark scores while exhibiting markedly different token generation behavior, producing short reasoning on some tasks and excessively long reasoning traces on others.

Second, conventional pruning applies the same reduced expert pool to both prefill and decode, sacrificing model quality during compute-bound prefill for little throughput benefit. This observation motivates SlimWise’s design: retain full-model prefill and prune only decode. We hypothesize that preserving full-model prompt representations during prefill can help a pruned decoder recover accuracy without restoring its removed experts.

\begin{figure}[t]
\centering
\begin{minipage}[b]{0.64\linewidth}
  \centering
  \includegraphics[width=\linewidth]{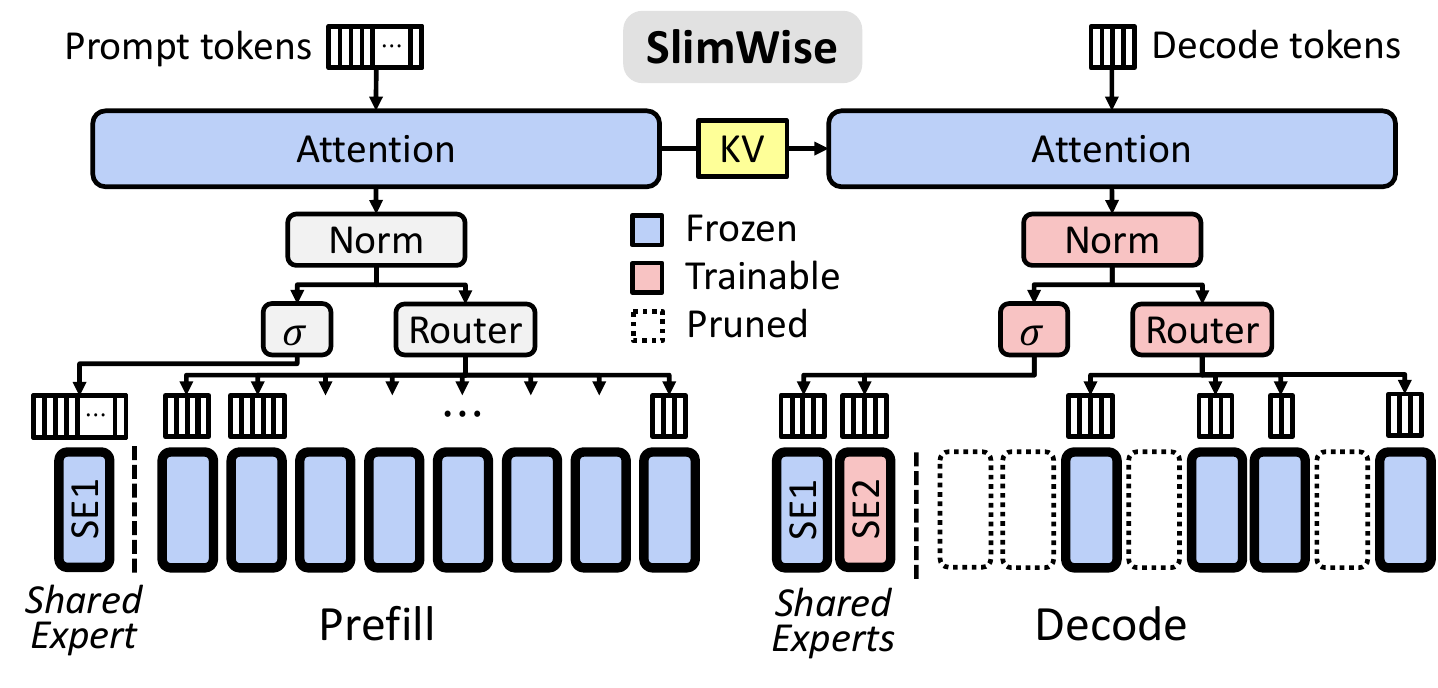}\\[-2pt]
  {\small (a)}
\end{minipage}\hfill
\begin{minipage}[b]{0.34\linewidth}
  \centering
  \includegraphics[width=\linewidth]{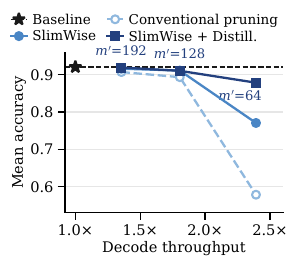}\\[-2pt]
  {\small (b)}
\end{minipage}
\caption{%
(a) Overview of SlimWise. Prefill uses the full expert pool, while decode masks out half of the expert pool with pruning but keeps the number of active experts per token, $k$, unchanged, directly reusing the KV cache generated during prefill.
(b) Accuracy–throughput trade-off under PD-disaggregated serving. Mean accuracy across five benchmarks (HE+, MBPP+, GSM8K, MATH-500, and BFCL) is plotted against decode throughput per GPU, normalized to the full-model baseline without pruning, at a per-user generation SLO of 50 tokens/s. Results use Qwen3.6-35B-A3B with REAP-selected expert sets, 1k input tokens, and 8k output tokens.
}
\label{fig:overview}
\end{figure}

SlimWise is a serving framework that allocates expert capacity separately to the two inference phases, combining full-model prefill with a pruned decoder that directly reuses the KV cache generated during prefill (Figure~\ref{fig:overview}a). Because expert pruning preserves the attention architecture and KV cache format, this \emph{KV cache handoff} requires no KV cache conversion. SlimWise naturally fits prefill–decode (PD) disaggregation~\citep{zhong2024distserve, patel2024splitwise, qin2026mooncake}, extending the flexibility to use different hardware and parallelism configurations across phases to the allocation of model capacity itself. It also supports PD-colocated serving: expert pruning can be expressed as router masking, allowing an engine to retain the full expert pool and apply the original router during prefill and a masked router during decode. We refer to this mechanism as \emph{phase-aware masking}.

Our central finding is that the KV cache handoff recovers a substantial portion of pruning-induced accuracy loss without additional training, narrowing the gap to the full model in multiple evaluated settings. Interestingly, its benefits also extend to generation behavior. Pruned prefill can prematurely shorten reasoning, whereas pruned decode can prolong generation, sometimes reaching the token budget without terminating. We observe that our KV cache handoff helps alleviate the former distortion but does not fully resolve the latter. We therefore introduce a low-cost distillation stage that trains the pruned decoder to continue from full-model KV caches. Updating only a small subset of parameters further improves accuracy and brings generation lengths closer to those of the full model.

We implement SlimWise in vLLM for both PD-disaggregated and PD-colocated serving. Decode throughput improves through two mechanisms. First, reducing the expert pool shrinks the set of distinct experts that each decode step must load, lowering expert-weight traffic and shortening memory-bound execution. Second, under PD disaggregation, the decode instance stores only the retained expert weights, freeing memory for additional KV cache capacity and larger batches, a benefit unavailable to runtime expert-selection methods that retain all experts~\citep{gupta2024lynx,wu2026exfold,oncescu2025opportunistic}. PD-colocated serving with SlimWise retains the full pool for prefill but still benefits from reduced expert-weight traffic during decode via our phase-aware masking. Figure~\ref{fig:overview}b illustrates the resulting accuracy–throughput trade-off. 
Under PD-disaggregation with a per-user generation target of 50 tokens/s, a decoder retaining one quarter of the experts achieves 2.39$\times$ the full model’s decode throughput. After distillation, its mean accuracy remains within a few percentage points of the full model’s, whereas conventional pruning at the same ratio incurs substantially greater accuracy loss. 
Across the evaluated per-user speed targets, SlimWise achieves up to 1.81$\times$ decode throughput at 50$\%$ expert pruning with minimal accuracy loss.

Our contributions are as follows:
\begin{itemize}
    \item We introduce SlimWise, which combines full-model prefill with pruned decode over a shared KV cache. This training-free KV cache handoff recovers a substantial portion of the accuracy lost to conventional pruning across various pruning criteria and MoE backbones.

    \item We also demonstrate that benchmark accuracy can conceal pruning-induced distortions in reasoning length and completion behavior. SlimWise introduces a low-cost distillation stage to train the pruned decoder on the KV cache it will inherit at deployment, improving accuracy and reducing generation-length distortions.
  
    \item We implement SlimWise in vLLM for both PD-disaggregated and PD-colocated serving, using phase-aware router masking for the latter. At 50\% expert pruning, SlimWise improves decode throughput by up to 1.81$\times$ with minimal accuracy loss.
\end{itemize}

\section{Background}
\label{sec:background}

\subsection{Batched MoE Inference}
\label{sec:bg-batched}

Despite per-token MoE sparsity, each decode step reads the weights of all distinct experts selected across the batch from the GPU's high-bandwidth memory (HBM).
Let $m$ denote the number of routed experts in a layer, $k$ active experts per token, and $B$ the decode batch size. Assuming uniform routing and independent expert selections across tokens, the expected number of distinct experts accessed is
\begin{equation}
  E_{\text{distinct}} = m\left[1-\left(1-\frac{k}{m}\right)^{B}\right].
  \label{eq:distinct}
\end{equation}
As $B$ grows, $E_{distinct}$ approaches the full expert pool $m$, so per-token sparsity does not guarantee per-batch sparsity. This distinction matters as models adopt larger expert pools while keeping the number of active experts per token fixed: Qwen3.6-35B-A3B activates 8 of 256 experts per token, versus 8 of 128 in Qwen3-30B-A3B, halving $k$/$m$ despite comparable total parameter counts.

\begin{figure}[t]
\centering
\includegraphics[width=0.48\linewidth]{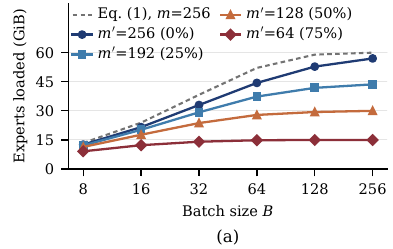}
\hfill
\includegraphics[width=0.48\linewidth]{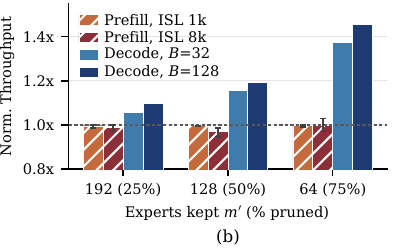}
\caption{%
Batched decoding reduces effective MoE sparsity, while expert pruning improves decode throughput with little effect on prefill. (a) Expert-weight volume accessed per decode step across batch sizes and pruning ratios. (b) Prefill and decode throughput normalized to the unpruned model ($m$=$m'$=256). Results use Qwen3.6-35B-A3B ($m=256$, $k=8$, 40 layers) in BF16 on $2\times$ A100-80GB GPUs, with nested REAP-selected expert sets applied as router masks. Percentages indicate the fraction of routed experts pruned.
}
\label{fig:batched-moe}
\end{figure}

Equation~\ref{eq:distinct}  assumes uniform routing, but pretrained MoE models often route tokens unevenly across experts. This skew concentrates tokens on a subset of experts, reducing the number of distinct experts accessed relative to the uniform-routing prediction. As our measurements show, however, skew delays saturation without preventing $E_{\mathrm{distinct}}$ from approaching the full expert pool $m$ as the batch size increases. Figure~\ref{fig:batched-moe}a measures the expert-weight volume accessed per decode step by Qwen3.6-35B-A3B on prompts sampled from \texttt{mlabonne/open-perfectblend}~\citep{labonne2024openperfectblend}. At a batch size of 64, each step accesses approximately 75\% of the 60 GiB expert pool; at the largest evaluated batch size, it accesses nearly the entire pool. 
Pruning lowers this saturation ceiling: retaining $m'$ of the original $m$ experts reduces the expert-weight volume accessed per step at saturation to approximately $m'/m$ of the original 60 GiB pool.


Accessing nearly every expert, however, does not imply enough computation to amortize weight-loading overhead.
As the accessed set approaches the full expert pool, each expert processes approximately $B\cdot(k/m)$ tokens per step on average—fewer than ten even at the largest batch size evaluated with Qwen3.6-35B-A3B. 
Larger batches increase weight reuse, but KV cache capacity and per-token latency requirements limit practical batch sizes, leaving decode dominated by expert-weight movement. 
Prefill offers substantially greater weight reuse by processing many prompt tokens in parallel. 
In this compute-bound regime, shrinking the expert pool while preserving $k$ leaves the dominant routed-expert computation per token unchanged, yielding little throughput benefit. 
Figure~\ref{fig:batched-moe}b illustrates this asymmetry: across the evaluated pruning ratios, prefill throughput stays within a few percent of the unpruned model with no consistent improvement, while decode throughput improves by up to approximately 1.5$\times$. 
These results motivate allocating expert capacity separately to the two phases.

\subsection{Prefill--Decode Disaggregation}
\label{sec:bg-pd}

Prefill and decode are typically compute- and memory-bound, respectively, with service level objectives (SLOs) on time-to-first-token (TTFT) for prefill and time-per-output-token (TPOT) for decode. Long prefill operations can degrade TPOT in PD-colocated serving. PD-disaggregated serving places the phases on separate instances and transfers the prefill-generated KV cache to decode~\citep{zhong2024distserve, patel2024splitwise, qin2026mooncake}, reducing interference and enabling phase-specific hardware and parallelism configurations.

The KV cache links prefill and decode, carrying keys and values for standard attention and, in hybrid models such as Qwen3.6, recurrent states for linear attention. Expert pruning may change these states’ values but preserves the attention architecture and KV cache format. A pruned decoder can therefore directly consume the KV cache produced by the full model without conversion, allowing the two phases to use different expert pools while communicating through the same KV cache interface.

\subsection{Reducing Expert Traffic}
\label{sec:bg-reduction}

Existing approaches reduce expert-weight traffic through runtime expert selection or model compression. Lynx~\citep{gupta2024lynx} and Opportunistic Expert Activation (OEA)~\citep{oncescu2025opportunistic} reroute tokens to already-active experts, while ExFold~\citep{wu2026exfold} folds excluded experts into retained ones using calibrated scalar projectors. 
Note that these methods retain the full expert pool in memory.
In contrast, pruning~\citep{jaiswal2025finding, xie2024moe} and merging~\citep{li2024merge, chen2025hcsmoe} reduce the pool itself. 
We focus on pruning, which recent evaluations favor over merging on generative tasks~\citep{lasby2026reap}.

We consider three state-of-the-art pruning criteria computed from calibration rollouts: routing mass aggregates gate probabilities; Expert Activation Norm (EAN) measures expert output norms~\citep{jaiswal2025finding}; and Router-weighted Expert Activation Pruning (REAP) weights these norms by gate probabilities~\citep{lasby2026reap}. Each criterion ranks experts within each layer to select a fixed retained set, keeping the number of active experts per token, $k$, unchanged.

\section{Methodology}
\label{sec:method}

\subsection{SlimWise Serving Framework}
\label{sec:framework}

The observations in Section~\ref{sec:background} motivate allocating expert capacity separately to prefill and decode. Expert pruning reduces the weight traffic that limits decode performance, while offering little throughput benefit during compute-bound prefill. Moreover, pruning preserves the KV cache format, allowing a pruned decoder to reuse the states produced by full-model prefill.
Let $K_\ell \subseteq \{1,\dots,m\}$ denote the unpruned, original expert set for layer $\ell$, with $|K_\ell|=m \geq k$. We construct $K_\ell'$ by ranking the layer’s experts using an importance criterion and retaining the top $m' \leq m$.
SlimWise runs prefill with the full expert pool and decode using only $K_\ell'$. The pruned decoder directly inherits the full model’s KV cache without conversion because the KV cache structure is independent of $m'$.

The pruned decoder admits two equivalent implementations: a separate, pruned model checkpoint containing only the retained experts and their corresponding router entries ($K_\ell'$), or the full expert pool with a mask applied to the router.
Given router logits $z \in \mathbb{R}^{m}$, we define the masked logits as
\begin{equation}
  \tilde z_i =
  \begin{cases}
    z_i & i \in K_\ell', \\
    \tau & \text{otherwise},
  \end{cases}
  \label{eq:mask}
\end{equation}
where $\tau$ is a large negative value to exclude masked experts from top-$k$ selection. The masked router selects the top $k$ within $K_\ell'$, exactly what the pruned model selects. 
Renormalization runs over the selected $k$ alone, so the gate each surviving expert receives does not depend on whether the removed experts are present.
This equivalence supports both PD-disaggregated and PD-colocated serving. Under PD-disaggregated serving, the decode instance loads the pruned checkpoint, allowing memory previously occupied by removed experts to be reallocated to the KV cache.
Runtime expert-selection methods that retain the full expert pool do not provide this memory saving~\citep{gupta2024lynx, oncescu2025opportunistic, wu2026exfold}.
Under PD-colocated serving, the engine retains the full expert pool and applies phase-aware router masking: prefill uses the original router, while decode uses the masked router in Equation~\ref{eq:mask}. The full model, conventional pruning, and training-free SlimWise can therefore share a single implementation of the MoE serving framework, with masking applied to neither phase (full), both phases (conventional), or decode alone (SlimWise), respectively.

\subsection{Benefits and Limitations of SlimWise KV Cache Handoff}
\label{sec:handoff}

\begin{figure}[t]
  \centering
  \includegraphics[width=0.48\textwidth]{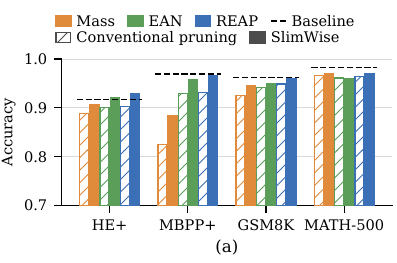}\hfill
  \includegraphics[width=0.48\textwidth]{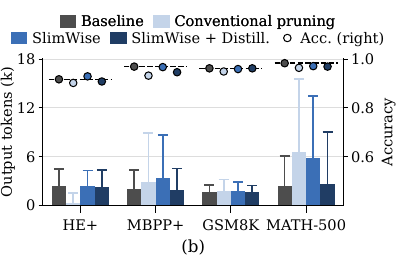}%
  \caption{
  Accuracy and generation length at 50\% expert pruning ($m'$=128) on Qwen3.6-35B-A3B, using sampled decoding ($T{=}1.0$, top-$p{=}0.95$, top-$k{=}20$) with thinking enabled. \emph{Conventional pruning} uses the pruned model for both prefill and decode; \emph{SlimWise} preserves full-model prefill while using the pruned decoder. (a) Accuracy without additional training across three pruning criteria. (b) Output-length distributions and accuracy under REAP pruning, including SlimWise with distillation. Bars show median output lengths ($p_{50}$), whiskers indicate the 90th percentile ($p_{90}$), and circles show accuracy on the right axis. Dashed lines mark full-model accuracy in both panels.
  }
  \label{fig:handoff-sampled}
\end{figure}

We first evaluate whether SlimWise’s KV cache handoff can recover accuracy lost to expert pruning without additional training. Figure~\ref{fig:handoff-sampled}a compares conventional pruning with SlimWise on Qwen3.6-35B-A3B after removing half of the routed experts using each of three state-of-the-art pruning criteria. Although the extent of accuracy degradation varies across criteria and benchmarks, SlimWise substantially narrows the gap to the full-model baseline in several of the most affected settings. These gains occur across all three pruning criteria without changing the retained expert sets, demonstrating that preserving full-model prefill can help recover accuracy while using the same pruned decoder.

Benchmark scores alone, however, do not fully capture the effect pruning has on model quality. Figure~\ref{fig:handoff-sampled}b compares accuracy with output-length distributions under the REAP criterion. On HumanEval+ (HE+), conventional pruning substantially reduces the median output length from 2.3k to 0.3k tokens, yet accuracy decreases only modestly. On MATH-500, the distortion runs in the opposite direction: the median increases from 2.4k to 6.6k tokens, and the 90th percentile also grows substantially, while accuracy remains close to the baseline. Overall, these results show that benchmark scores alone do not fully capture pruning’s effects on token generation, highlighting the need to evaluate not just model accuracy but also the model's token generation behavior.
Excessively long reasoning traces have also been reported under post-training quantization~\citep{lotfi2026quantized}, highlighting a related concern for reasoning efficiency.

SlimWise's KV cache handoff helps alleviate these distortions but does not eliminate them completely: it restores the HumanEval+ median output length to approximately 2.3k tokens, close to the baseline, but on MATH-500, the median remains approximately 5.8k tokens, more than twice the baseline’s. On MBPP+, the KV cache handoff further increases the already elevated median output length. Preserving full-model prefill can thus restore generation behavior in some settings while leaving excessive generation unresolved in others. It is worth emphasizing that such behavior is not an artifact of our sampling procedure. Under greedy decoding (Figure~\ref{fig:handoff} in the Appendix), generations run to the token budget without terminating naturally, both with and without the KV cache handoff. These results point to the pruned decoder itself as the source of the remaining distortions, motivating the distillation stage described in Section~\ref{sec:distill}. Its effect on generation behavior is also shown in Figure~\ref{fig:handoff-sampled}b.

\subsection{Recovering the Remaining Loss With a Low-Cost Distillation}
\label{sec:distill}

To address residual accuracy loss and generation-length distortions, we fine-tune the pruned decoder through low-cost distillation from the full model.
Unlike prior work that retrains pruned students on billions of tokens~\citep{basant2025nvidia, tang2026slimqwen, kim2026pruning}, we merely use 100M tokens and update only a small subset of parameters.
Our procedure also aligns the student’s training conditions with SlimWise’s unique deployment setting: while conventional distillation conditions the student on its own KV states throughout the sequence, SlimWise trains it to continue from teacher-generated KV caches, reproducing the KV cache handoff from full-model prefill to pruned decode.

For each training sequence of length $L$, we sample a split point $s$ uniformly from $[0, L)$. The teacher (the full model) processes the prefix $[0, s)$ to produce a KV cache, which is then handed off to the student (the pruned decoder) for processing the remaining tokens. 
Let $C_T$ and $C_S$ denote the KV cache states produced by the teacher and student, respectively. 
At each position $t \ge s$, the student predicts according to $p_S\big(y_t \mid C_T(x_{0:s}),\, C_S(x_{s:t})\big)$, where $x_{a:b}$ denotes the tokens in $[a, b)$. The student KV cache states are computed by continuing from the teacher’s prefix cache.
Conventional distillation corresponds to $s=0$, for which the student produces all of its own KV cache and predicts with $p_S\big(y_t \mid C_S(x_{0:t})\big)$. At deployment, the KV cache handoff occurs at the end of the prompt, so sampling $s$ uniformly from $[0, L)$ exposes the student to a range of prefix lengths, reflecting the variation in prompt lengths across requests.

We compute the loss over positions in $[s, L)$, combining cross-entropy ($\mathcal{L}_{\text{CE}}$) with distillation ($\mathcal{L}_{\text{KD}}$) against the teacher's distribution $p_T\big(y_t \mid C_T(x_{0:t})\big)$:
\begin{equation}
  \mathcal{L} = \alpha \, \mathcal{L}_{\text{CE}} + (1-\alpha)\, \mathcal{L}_{\text{KD}},
  \qquad \alpha = 0.1 .
  \label{eq:loss}
\end{equation}
The distillation term $\mathcal{L}_{\text{KD}}$ is a KL divergence over the teacher’s top-$K$ tokens, with $K=64$, and a residual bucket that aggregates the probability mass of all remaining tokens. 
This formulation preserves the teacher’s original probabilities for the selected tokens while retaining its aggregate probability mass outside the top-$K$ set, encouraging the student to match both.

To keep the cost of SlimWise's distillation inexpensive, we restrict training to the always-on components like the shared expert, routers, and normalization layers, leaving the routed expert weights and attention parameters frozen.
Together they account for 0.42\% of the parameters in Qwen3.6-35B-A3B and 2.12\% in Gemma 4-26B-A4B.
We share all frozen tensors between the teacher and student, allowing both models to reside on the same GPU without storing two complete model copies. Because even small router updates can change token-to-expert assignments, we set the router learning rate to one-tenth that of the other trainable parameters.

We freeze the original shared expert (SE1 in Figure~\ref{fig:overview}a) and add a second expert of the same intermediate width (SE2), with its down projection initialized to zero. This preserves the pruned block’s output at initialization while allowing SE2 to learn an additive correction during distillation, analogous to the zero-output initialization used in LoRA~\citep{hu2021lora} and ControlNet~\citep{zhang2023adding}. Because both experts share the same sigmoid gate, their concatenated outputs can be represented by a single expert with twice the intermediate width. We export this combined form as a standard checkpoint, allowing deployment without adding custom serving operations.

\section{Evaluation Results}
\label{sec:results}

\subsection{Experimental Setup}
\label{sec:setting}

We evaluate SlimWise on Qwen3.6-35B-A3B and Gemma 4-26B-A4B-it~\citep{team2026gemma}.
For each model, we collect expert statistics in a single pass over the REAP calibration mixture~\citep{lasby2026reap} and use these statistics to rank experts under the three criteria described in Section~\ref{sec:bg-reduction}. Each configuration retains the same number of experts in every layer.
We evaluate math on GSM8K~\citep{cobbe2021training} and MATH-500~\citep{lightman2024let}, coding on HumanEval+ and MBPP+~\citep{liu2023your}, and tool use on BFCLv4~\citep{patil2025berkeley}. All benchmarks use sampled decoding with thinking enabled and the generation budget is 32{,}768 tokens per response. We report mean accuracy across three runs. Distillation uses 64k training sequences of at most 12k tokens and updates the parameters described in Section~\ref{sec:distill} for 8,000 steps. Each run takes approximately six hours on four B200 GPUs. Appendix~\ref{app:setting} provides further details on calibration, benchmarks, and distillation.

\subsection{Accuracy Evaluation}

\begin{table}[t]
\centering
\small
\caption{Accuracy and output length at 50\% expert pruning across two MoE backbones and three pruning criteria. All results use sampled decoding ($T$=1.0, top-$p$=0.95; top-$k$=20 for Qwen3.6, 64 for Gemma~4) with thinking enabled. Accuracy is averaged over three runs;
the parenthesized value is the median output length over all responses pooled across runs. Bold marks the highest accuracy for each model, pruning criterion, and benchmark; underlining marks the highest accuracy across all pruned configurations for each model and benchmark.}
\setlength{\tabcolsep}{3pt}
\begin{tabular}{llcccccc}
\toprule
 & & \multicolumn{2}{c}{Math} & \multicolumn{2}{c}{Coding} & \multicolumn{2}{c}{Tool-Use (BFCLv4)} \\
\cmidrule(lr){3-4} \cmidrule(lr){5-6} \cmidrule(lr){7-8}
Model & Method & GSM8K & MATH-500 & HE+ & MBPP+ & Non-Live & Live \\
\midrule
\multirow{10}{*}{\shortstack[c]{Qwen3.6\\(35B-A3B)}}
 & Baseline                     & 0.962 \dk{1.7k} & 0.983 \dk{2.4k} & 0.917 \dk{2.3k} & 0.969 \dk{1.9k} & 0.884 \dk{663} & 0.811 \dk{567} \\
\cmidrule(l){2-8}
 & Mass                         & 0.925 \dk{1.7k} & 0.966 \dk{2.6k} & 0.888 \dk{3.7k} & 0.825 \dk{5.3k} & 0.793 \dk{484} & 0.680 \dk{526} \\
 & \quad SlimWise               & \textbf{0.946} \dk{1.6k} & \underline{\textbf{0.971}} \dk{2.6k} & 0.907 \dk{2.5k} & 0.883 \dk{2.9k} & 0.803 \dk{532} & 0.740 \dk{563} \\
 & \quad + Distill.             & \textbf{0.946} \dk{1.6k} & 0.967 \dk{2.4k} & \textbf{0.921} \dk{2.4k} & \textbf{0.919} \dk{2.1k} & \textbf{0.872} \dk{611} & \textbf{0.792} \dk{542} \\
\cmidrule(l){2-8}
 & EAN                          & 0.941 \dk{2.6k} & 0.961 \dk{8.5k} & 0.900 \dk{1.1k} & 0.930 \dk{1.0k} & 0.829 \dk{387} & 0.730 \dk{473} \\
 & \quad SlimWise               & 0.949 \dk{1.9k} & 0.959 \dk{6.5k} & \textbf{0.921} \dk{2.0k} & \textbf{0.957} \dk{2.4k} & 0.842 \dk{544} & 0.766 \dk{588} \\
 & \quad + Distill.             & \textbf{0.960} \dk{1.6k} & \textbf{0.967} \dk{2.7k} & 0.917 \dk{2.3k} & 0.936 \dk{2.2k} & \underline{\textbf{0.877}} \dk{536} & \underline{\textbf{0.804}} \dk{523} \\
\cmidrule(l){2-8}
 & REAP                         & 0.949 \dk{1.7k} & 0.964 \dk{6.6k} & 0.902 \dk{0.3k} & 0.932 \dk{2.8k} & 0.857 \dk{355} & 0.759 \dk{469} \\
 & \quad SlimWise               & 0.959 \dk{1.7k} & \underline{\textbf{0.971}} \dk{5.8k} & \underline{\textbf{0.929}} \dk{2.3k} & \underline{\textbf{0.967}} \dk{3.4k} & 0.859 \dk{599} & 0.784 \dk{575} \\
 & \quad + Distill.             & \underline{\textbf{0.962}} \dk{1.6k} & 0.969 \dk{2.6k} & 0.909 \dk{2.3k} & 0.946 \dk{1.9k} & \underline{\textbf{0.877}} \dk{575} & \textbf{0.802} \dk{522} \\
\midrule
\multirow{10}{*}{\shortstack[c]{Gemma~4\\(26B-A4B)}}
 & Baseline                     & 0.971 \dk{0.8k} & 0.962 \dk{1.6k} & 0.949 \dk{1.8k} & 0.979 \dk{0.9k} & 0.831 \dk{215} & 0.807 \dk{255} \\
\cmidrule(l){2-8}
 & Mass                         & 0.914 \dk{1.3k} & 0.896 \dk{2.3k} & 0.917 \dk{1.8k} & 0.945 \dk{0.9k} & 0.824 \dk{219} & 0.690 \dk{312} \\
 & \quad SlimWise               & 0.938 \dk{0.9k} & 0.906 \dk{2.1k} & 0.907 \dk{2.0k} & 0.954 \dk{0.9k} & \underline{\textbf{0.825}} \dk{224} & 0.722 \dk{288} \\
 & \quad + Distill.             & \textbf{0.955} \dk{0.8k} & \underline{\textbf{0.945}} \dk{1.8k} & \underline{\textbf{0.947}} \dk{1.7k} & \underline{\textbf{0.967}} \dk{0.9k} & 0.805 \dk{211} & \textbf{0.773} \dk{253} \\
\cmidrule(l){2-8}
 & EAN                          & 0.816 \dk{0.1k} & 0.549 \dk{0.4k} & 0.246 \dk{2.1k} & 0.343 \dk{1.3k} & 0.007 \dk{32.8k} & 0.012 \dk{32.8k} \\
 & \quad SlimWise               & 0.173 \dk{32.8k} & 0.445 \dk{32.8k} & 0.494 \dk{32.8k} & 0.608 \dk{32.8k} & 0.052 \dk{32.8k} & 0.063 \dk{32.8k} \\
 & \quad + Distill.             & \underline{\textbf{0.965}} \dk{0.6k} & \textbf{0.843} \dk{1.4k} & \textbf{0.917} \dk{1.6k} & \textbf{0.929} \dk{0.8k} & \textbf{0.734} \dk{170} & \textbf{0.721} \dk{157} \\
\cmidrule(l){2-8}
 & REAP                         & 0.744 \dk{0.1k} & 0.763 \dk{0.3k} & 0.762 \dk{1.0k} & 0.804 \dk{0.9k} & 0.388 \dk{58} & 0.258 \dk{333} \\
 & \quad SlimWise               & 0.845 \dk{0.7k} & 0.736 \dk{2.8k} & 0.858 \dk{32.8k} & 0.841 \dk{32.8k} & 0.662 \dk{106} & 0.663 \dk{191} \\
 & \quad + Distill.             & \textbf{0.963} \dk{0.7k} & \textbf{0.921} \dk{1.3k} & \textbf{0.919} \dk{1.7k} & \textbf{0.958} \dk{0.8k} & \textbf{0.809} \dk{181} & \underline{\textbf{0.797}} \dk{192} \\
\bottomrule
\end{tabular}
\label{tab:qwen-gemma-sampled-50}
\end{table}

Table~\ref{tab:qwen-gemma-sampled-50} reports accuracy and median output length at 50\% expert pruning across both model backbones and all three pruning criteria. The results extend the observations in Section~\ref{sec:handoff}: pruning affects benchmarks differently depending on the criterion, and preserving full-model prefill recovers accuracy in many settings without additional training. The recovery is particularly pronounced for Gemma 4’s tool use under REAP pruning, significantly increasing BFCL accuracy while using the same pruned decoder. However, it also illustrates the limitations of SlimWise's KV cache handoff.
Under EAN pruning, the median output length reaches the generation budget on all math and coding benchmarks after the KV cache handoff, rendering the accuracy of SlimWise without distillation on GSM8K and MATH-500 to fall below that of conventional pruning. Adding distillation on top of KV cache handoff substantially improves accuracy and brings median output lengths closer to the baseline in these disrupted Gemma 4 settings.
Appendix~\ref{app:greedy} reports the same evaluation under greedy decoding, where both directions of the length distortion appear in sharper form, and Table~\ref{tab:sampled-std} in the Appendix reports the standard deviation across runs.

\begin{table}[]
\centering
\small
\caption{
Accuracy and output length across retained expert counts $m'$ on Qwen3.6-35B-A3B with REAP pruning, using sampled decoding with thinking enabled. Accuracy (Acc) is averaged over three runs; $p_{50}$ and $p_{90}$ denote the median and 90th-percentile output lengths, respectively, in thousands of tokens, pooled across runs. $\dagger$ denotes students distilled using their own prefill KV caches ($s{=}0$ in Section~\ref{sec:distill}). Bold marks the highest accuracy for each pruning level and benchmark.
}
\setlength{\tabcolsep}{3.2pt}
\begin{tabular}{cl c@{\hspace{8pt}}cc@{\hspace{12pt}} c@{\hspace{8pt}}cc@{\hspace{12pt}} c@{\hspace{8pt}}cc@{\hspace{12pt}} c@{\hspace{8pt}}cc}
\toprule
 & & \multicolumn{6}{c}{Math} & \multicolumn{6}{c}{Coding} \\
\cmidrule(lr){3-8} \cmidrule(lr){9-14}
 & & \multicolumn{3}{c}{GSM8K} & \multicolumn{3}{c}{MATH-500} & \multicolumn{3}{c}{HE+} & \multicolumn{3}{c}{MBPP+} \\
\cmidrule(lr){3-5} \cmidrule(lr){6-8} \cmidrule(lr){9-11} \cmidrule(lr){12-14}
$m'$ & Method & Acc & $p_{50}$ & $p_{90}$ & Acc & $p_{50}$ & $p_{90}$ & Acc & $p_{50}$ & $p_{90}$ & Acc & $p_{50}$ & $p_{90}$ \\
\midrule
256 & Baseline      &  0.962 &   1.7k &   2.5k &  0.983 &   2.4k &   6.1k &  0.917 &   2.3k &   4.5k &  0.969 &   1.9k &   4.4k \\
\midrule
192 & REAP                   &  0.956 &   1.8k &   2.7k &  \textbf{0.983} &   2.5k &   6.3k &  0.915 &   2.3k &   4.6k &  0.969 &   2.5k &   8.1k \\
    & \quad SlimWise         &  \textbf{0.959} &   1.7k &   2.7k &  0.978 &   2.5k &   6.2k &  \textbf{0.919} &   2.4k &   4.5k &  0.959 &   2.2k &   6.7k \\
    & \quad + Distill.       &  0.954 &   1.6k &   2.5k &  0.981 &   2.5k &   6.1k &  0.915 &   2.3k &   4.4k &  \textbf{0.974} &   1.9k &   4.2k \\
\midrule
64  & REAP                   &  0.742 &   0.3k &   4.9k &  0.831 &   0.7k &  11.7k &  0.331 &   0.2k &   1.8k &  0.524 &   0.6k &   7.7k \\
    & + Distill.$^\dagger$   &  0.886 &   1.6k &   6.7k &  0.912 &   7.2k &  26.2k &  0.805 &   1.7k &   4.7k &  0.783 &   0.2k &   0.7k \\
    & \quad SlimWise         &  0.867 &   1.1k &   4.7k &  0.935 &   4.7k &  32.8k &  0.856 &   1.2k &   4.6k &  0.844 &   1.0k &   5.6k \\
    & \quad + Distill.$^\dagger$ &  0.938 &   1.6k &   2.6k &  0.928 &   3.1k &  22.4k &  0.890 &   2.4k &   4.8k &  \textbf{0.929} &   2.2k &   8.1k \\
    & \quad + Distill.       &  \textbf{0.959} &   1.4k &   2.6k &  \textbf{0.949} &   2.8k &  14.0k &  \textbf{0.894} &   1.3k &   3.4k &  0.890 &   0.8k &   3.5k \\
\bottomrule
\end{tabular}
\label{tab:qwen-ratio-sampled}
\end{table}

Table~\ref{tab:qwen-ratio-sampled} examines different numbers of retained experts under REAP pruning on Qwen3.6-35B-A3B. At $m'{=}192$ (25\% pruning), conventional pruning largely preserves accuracy, leaving little loss for our proposal to recover.
Nevertheless, conventional pruning increases the MBPP+ 90th-percentile output length from 4.4k to 8.1k tokens; SlimWise reduces it to 6.7k, and distillation further reduces it to 4.2k.
At $m'{=}64$ (75\% pruning), conventional pruning substantially degrades accuracy and shortens median responses across all four benchmarks. SlimWise's KV cache handoff recovers a substantial share of the accuracy loss, but the MATH-500 90th-percentile output length reaches the 32.8k-token budget. Adding distillation further improves accuracy on all four benchmarks and reduces this tail to 14.0k tokens, although it remains above the baseline’s 6.1k. 
The phase-swap experiments in Appendix~\ref{app:swap} suggest that pruned prefill (`Reverse' in Table~\ref{tab:swap}) contributes to shortened coding responses, while pruned decode (`SlimWise') contributes to prolonged generation on MATH-500.

Table~\ref{tab:qwen-ratio-sampled} also shows the effect of distilling the student using the inherited teacher KV cache. 
Rows marked $\dagger$ use students distilled entirely on their own prefill cached states ($s{=}0$ in Section~\ref{sec:distill}).
At 75\% pruning, giving the same conventionally distilled student the full model's cache at inference (SlimWise+Distill.$^\dagger$) improves accuracy on all four benchmarks and reduces the 90th-percentile output lengths on GSM8K and MATH-500, demonstrating a benefit from the handoff beyond additional training alone.
Distilling the student on the inherited teacher cache (SlimWise+Distill.) further improves accuracy on three of the four benchmarks and reduces the 90th-percentile output lengths on MATH-500, HumanEval+, and MBPP+.
Note that on MBPP+, this reduction involves a trade-off: accuracy decreases from 0.929 to 0.890, while the 90th-percentile output length falls from 8.1k to 3.5k tokens.

\subsection{Throughput Evaluation}
\label{sec:throughput}

\begin{figure}[t]
  \centering
  \includegraphics[width=\linewidth]{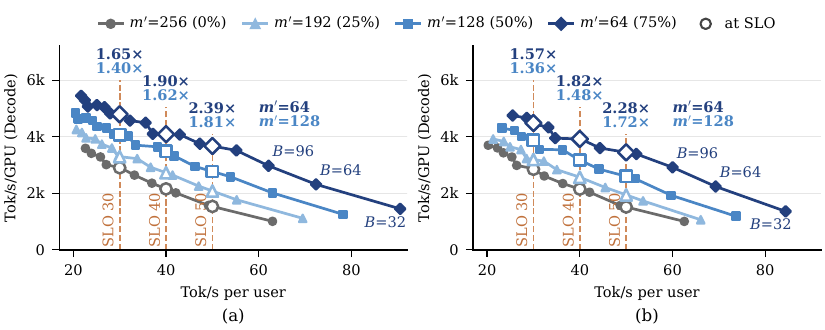}
  \caption{
  Per-GPU decode throughput versus per-user generation speed (SLO) on Qwen3.6-35B-A3B, using REAP-selected expert sets on two A100-80GB GPUs with 1k-token inputs and 8k-token outputs, over (a) PD-disaggregated serving and (b) PD-colocated serving. Each curve spans batch sizes up to the maximum supported by the available KV cache memory. Vertical lines indicate per-user SLOs, and hollow markers identify the corresponding operating points. Annotated numbers report throughput speedups over the full model at the same SLO for $m'{=}64$ and $m'{=}128$.
  }
  \label{fig:interactivity}
\end{figure}

Figure~\ref{fig:interactivity} plots per-GPU decode throughput (y-axis) against per-user generation speed (x-axis) across batch sizes for PD-disaggregated and PD-colocated serving. Per-user generation speed is the inverse of time per output token (TPOT), i.e., the SLO. The pruned decoders use the architecture of our distilled models, including a shared expert with twice its original intermediate width. Larger batches generally increase aggregate throughput at the expense of per-user generation speed. We therefore compare the maximum per-GPU throughput that satisfies each of three per-user generation-speed SLOs (30, 40, and 50 tokens/s) marked by the vertical lines.

Under PD-disaggregated serving (Figure~\ref{fig:interactivity}a), the decode instance loads a checkpoint containing only the retained experts. Pruning provides several benefits: it reduces expert-weight traffic, shortening decode steps, and frees memory for the KV cache. Shorter decode steps allow larger batches to satisfy the same per-user SLO, improving aggregate throughput. The additional KV cache capacity also increases the maximum feasible batch size, as reflected in the endpoints of the curves. Across the evaluated SLOs, SlimWise achieves 1.40–1.81$\times$ the full model’s decode throughput at 50\% pruning ($m'$=128) and 1.65–2.39$\times$ at 75\% pruning ($m'$=64), with larger relative gains under stricter SLOs.

Under PD-colocated serving (Figure~\ref{fig:interactivity}b), the full expert pool remains resident for prefill, while phase-aware router masking restricts decode to the retained experts. This configuration reduces expert-weight traffic during decode but does not free expert-weight memory for the KV cache, leaving all configurations with the same maximum batch size. 
SlimWise achieves slightly lower speedups than under PD-disaggregated serving (1.36--1.72$\times$ at 50\% pruning and 1.57--2.28$\times$ at 75\% pruning), partly due to router masking, which adds approximately 0.5\,ms per decode step, or 2.5--5\% of the step time.
Overall, SlimWise demonstrates its merits by improving throughput in both deployment configurations, with PD-disaggregated serving additionally benefiting from the smaller decoder memory footprint.

\section{Conclusion and Limitations}

SlimWise combines full-model prefill with pruned decode for efficient MoE serving. Without additional training, SlimWise's KV cache handoff recovers much of the accuracy lost to pruning in many evaluated settings. Our low-cost distillation stage further narrows the accuracy gap and mitigates generation-length distortions that benchmark scores alone can conceal. At 50\% expert pruning, SlimWise improves decode throughput by 1.4--1.8$\times$ while leaving prefill throughput unchanged. While these results are encouraging, SlimWise does come with two limitations. First, prefill cost remains unchanged, so the end-to-end speedup depends on the fraction of serving time spent in decode. Second, PD-colocated serving retains the full expert pool for prefill and therefore does not reduce the model-weight memory footprint.

\subsection*{AI Use Statement}

In this work, we used generative AI tools to implement methods and assist with translation and language editing.
We did not use generative AI tools to generate synthetic datasets, propose or refine hypotheses, help develop theoretical models or conceptual frameworks, design or provide feedback on research methodology or experiments, interpret results, formulate mathematical claims, or provide critical steps in proving mathematical claims. 
Dataset cleaning and reformatting, as well as qualitative and thematic data analysis, were not applicable to this work.
Additionally, we used generative AI tools to create or modify scientific figures or images, suggest experimental parameters, create or edit software code, and identify relevant literature.
We have reviewed all AI-assisted work. 
In particular, we manually verified the accuracy and relevance of AI-suggested related work by consulting the original papers.
We take responsibility for the final content of this work, including text, claims, and artifacts produced with the aid of generative AI.


\bibliography{iclr2027_conference}
\bibliographystyle{iclr2027_conference}

\appendix

\section{Appendix}



\subsection{Experimental Details}
\label{app:setting}

\paragraph{Pruning.}
The REAP calibration mixture~\citep{lasby2026reap} comprises six components, each containing 4,096 packed sequences of 2,048 tokens. For each model, we collect per-expert statistics in a single pass over this mixture and use them to rank experts within each layer under the three pruning criteria: routing mass, EAN, and REAP. The top $m'$ experts form the retained expert set. Each configuration retains the same number of experts in every layer while keeping the number of active experts per token, $k$, unchanged.


\paragraph{Benchmarks.}
We evaluate HumanEval+ and MBPP+ zero-shot, GSM8K with five-shot prompting on its first 500 problems, and MATH-500 with four-shot prompting. GSM8K and MATH-500 are graded using Math-Verify~\citep{mathverify}. Math and coding evaluations use the lm-evaluation-harness~\citep{evalharness}. BFCLv4 uses native function calling, and we report its Non-Live and Live AST aggregates. Since long chains of thought amplify run-to-run variation under continuous batching, every score is the mean of three runs.


\paragraph{Distillation.}

Training uses 64k sequences of at most 12k tokens each, totaling approximately 282M tokens for Qwen3.6 and 250M for Gemma~4, drawn from a corpus combining teacher-generated rollouts with thinking enabled on prompts from \texttt{open-perfectblend}~\citep{labonne2024openperfectblend} and multi-turn agentic traces from APIGen-MT~\citep{prabhakar2025apigen}, which account for 35\% of the corpus.
The student processes only the tokens after the split point $s$, totaling approximately 102M tokens for Qwen3.6 and 87M for Gemma~4.
We train for 8,000 steps using AdamW with a global batch size of eight and a learning rate of $10^{-4}$, reduced to $10^{-5}$ for routers. As described in Section~\ref{sec:distill}, training updates the shared expert in Qwen3.6 and the dense MLP in Gemma~4, together with routers and normalization layers. Sharing all frozen tensors allows the teacher and student to reside on the same GPU without storing two complete model copies. Each distillation run takes approximately six hours on four B200 GPUs.


\subsection{SlimWise with Greedy Decoding}
\label{app:greedy}

\begin{figure}[t]
  \centering
  \includegraphics[width=0.48\textwidth]{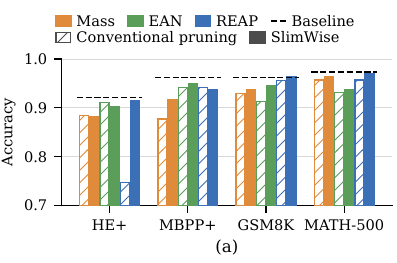}\hfill
  \includegraphics[width=0.48\textwidth]{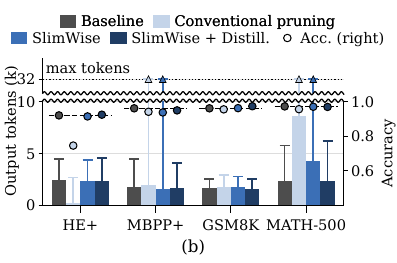}%
  \caption{
  Accuracy and generation length at 50\% expert pruning ($m'$=128) on Qwen3.6-35B-A3B, using greedy decoding with thinking enabled. \emph{Conventional pruning} uses the pruned model for both prefill and decode; \emph{SlimWise} preserves full-model prefill while using the pruned decoder. (a) Accuracy without additional training across three pruning criteria. (b) Output-length distributions (left-axis) and accuracy (right-axis) under REAP pruning, including SlimWise with distillation. Bars show median output lengths ($p_{50}$), whiskers indicate the 90th percentile ($p_{90}$), and circles show accuracy on the right axis. Dashed lines mark full-model accuracy in both panels.
  }
  \label{fig:handoff}
\end{figure}

Figure~\ref{fig:handoff} repeats the evaluation in Figure~\ref{fig:handoff-sampled} under greedy decoding, using the same retained expert sets, prompts, and generation budget. The generation-length distortions observed under sampled decoding (Section~\ref{sec:handoff}) persist and, in several settings, become more pronounced.
Under REAP pruning on HumanEval+, conventional pruning reduces accuracy from 0.921 to 0.746 and shortens the median output length from 2.4k to 0.3k tokens. SlimWise’s KV cache handoff recovers accuracy to 0.915 and restores the median length to 2.4k tokens. On MBPP+ and MATH-500, however, the 90th-percentile output length reaches the generation budget both with and without the KV cache handoff. Adding distillation reduces these percentiles below the budget and brings the output-length distributions closer to the baseline’s. These results confirm that the distortions described in Section~\ref{sec:handoff} are not solely artifacts of random token sampling.

\subsection{{Effects of Expert Pruning in Prefill and Decode}}
\label{app:swap}

SlimWise is motivated by the hypothesis that full-model prefill can mitigate pruning-induced accuracy loss even when paired with a pruned decoder.
Table~\ref{tab:swap} examines the effects of pruning each inference phase by comparing all four combinations of full and pruned models for prefill and decode: (1) full-prefill/full-decode (`Baseline'), (2) pruned-prefill/pruned-decode (`REAP'), (3) pruned-prefill/full-decode (`Reverse'), and (4) full-prefill/pruned-decode (both `SlimWise' and `SlimWise+Distill'). The `Reverse' configuration (pruned-prefill followed by full-model decode) complements SlimWise by testing whether restoring the full decoder can compensate for pruning during prefill. We evaluate Qwen3.6-35B-A3B at 50\% ($m'$=128) and 75\% pruning ($m'$=64), using the same REAP-selected expert sets across configurations at each pruning ratio.


The relative effects of pruning the two phases depend on the task. The benefit of preserving full-model prefill is particularly clear on HumanEval+. At 50\% pruning, the `Reverse' configuration achieves 0.890 accuracy with a median output length of 0.3k tokens, compared with the baseline’s 0.917 and 2.3k tokens. SlimWise achieves 0.929 accuracy and restores the median length to 2.3k tokens despite using the pruned decoder. At 75\% pruning, SlimWise also achieves higher accuracy than the `Reverse' configuration on both coding benchmarks, although neither configuration fully recovers baseline accuracy. The relative importance of the two phases nevertheless varies even within coding: at 50\% pruning, the `Reverse' configuration already brings MBPP+ accuracy and median output length close to the baseline.


On math benchmarks, restoring the full decoder provides greater accuracy recovery than preserving full-model prefill alone. `Reverse' retains accuracy close to the baseline on GSM8K and MATH-500 at 50\% pruning and outperforms training-free SlimWise on both benchmarks at 75\% pruning. The MATH-500 length distributions further suggest that pruned decode contributes to excessive generation. At 75\% pruning, Reverse produces a median output length of 2.2k tokens, close to the baseline’s 2.4k, whereas SlimWise produces a median of 4.7k and a 90th percentile reaching the generation budget. Distillation reduces SlimWise’s 90th-percentile length to 14.0k tokens, although it remains above the baseline’s 6.1k.

One might therefore ask whether distilling `Reverse' configuration's pruned prefill, analogous to distilling SlimWise’s pruned decoder, could recover accuracy competitive with the full-model baseline across tasks. This possibility, however, does not alter our design rationale: SlimWise aims to \emph{accelerate} MoE serving by exploiting the different performance characteristics of prefill and decode. Pruning compute-bound prefill while preserving the number of active experts per token offers little throughput benefit (Figure~\ref{fig:batched-moe}b), whereas retaining the full expert pool during decode leaves its primary performance bottleneck unaddressed. Distilling the pruned prefill could improve accuracy, but it would not reduce the decoder’s expert-weight footprint or directly alleviate its weight traffic. We therefore focus on distilling the pruned decoder, where quality recovery complements the serving benefits of reduced expert-weight traffic and, under PD disaggregation, increased KV cache capacity.



\begin{table}[h]
\centering
\small
\caption{Accuracy and output length across prefill–decode model configurations on Qwen3.6-35B-A3B with REAP-selected expert sets. REAP uses the pruned model for both phases; `Reverse' uses pruned prefill followed by full-model decode; and SlimWise uses full-model prefill followed by pruned decode. All results use sampled decoding ($T=1.0$, top-$p=0.95$, top-$k=20$) with thinking enabled and a 32,768-token generation budget.}
\setlength{\tabcolsep}{3.2pt}
\begin{tabular}{llcccccccccccc}
\toprule
& & \multicolumn{6}{c}{Math} & \multicolumn{6}{c}{Coding} \\
\cmidrule(lr){3-8} \cmidrule(lr){9-14}
& & \multicolumn{3}{c}{GSM8K} & \multicolumn{3}{c}{MATH-500} & \multicolumn{3}{c}{HE+} & \multicolumn{3}{c}{MBPP+} \\
\cmidrule(lr){3-5} \cmidrule(lr){6-8} \cmidrule(lr){9-11} \cmidrule(lr){12-14}
$m'$ & Method & Acc & $p_{50}$ & $p_{90}$ & Acc & $p_{50}$ & $p_{90}$ & Acc & $p_{50}$ & $p_{90}$ & Acc & $p_{50}$ & $p_{90}$ \\
\midrule
256 & Baseline          & 0.962 & 1.7k & 2.5k & 0.983 & 2.4k & 6.1k  & 0.917 & 2.3k & 4.5k & 0.969 & 1.9k & 4.4k \\
\midrule
\multirow{4}{*}{128}
& REAP              & 0.949 & 1.7k & 3.2k & 0.964 & 6.6k & 15.5k & 0.902 & 0.3k & 1.5k & 0.932 & 2.8k & 8.9k \\
& \quad \textbf{Reverse}     & 0.962 & 1.7k & 2.6k & 0.976 & 2.4k & 9.9k  & 0.890 & 0.3k & 2.6k & 0.966 & 2.0k & 4.1k \\
& \quad SlimWise    & 0.959 & 1.7k & 2.9k & 0.971 & 5.8k & 13.5k & 0.929 & 2.3k & 4.3k & 0.967 & 3.4k & 8.6k \\
& \quad + Distill.  & 0.962 & 1.6k & 2.4k & 0.969 & 2.6k & 9.0k  & 0.909 & 2.3k & 4.4k & 0.946 & 1.9k & 4.5k \\
\midrule
\multirow{4}{*}{64}
& REAP              & 0.742 & 0.3k & 4.9k & 0.831 & 0.7k & 11.7k & 0.331 & 0.2k & 1.8k & 0.524 & 0.6k & 7.7k \\
& \quad \textbf{Reverse}     & 0.915 & 0.4k & 2.7k & 0.965 & 2.2k & 13.0k & 0.521 & 0.5k & 3.0k & 0.634 & 0.5k & 2.2k \\
& \quad SlimWise    & 0.867 & 1.1k & 4.7k & 0.935 & 4.7k & 32.8k & 0.856 & 1.2k & 4.6k & 0.844 & 1.0k & 5.6k \\
& \quad + Distill.  & 0.959 & 1.4k & 2.6k & 0.949 & 2.8k & 14.0k & 0.894 & 1.3k & 3.4k & 0.890 & 0.8k & 3.5k \\
\bottomrule
\end{tabular}
\label{tab:swap}
\end{table}

\begin{table}[t]
\centering
\small
\caption{Accuracy at 50\% expert pruning across two MoE backbones and three pruning criteria. All evaluations use sampled decoding ($T=1.0$, top-$p=0.95$; top-$k=20$ for Qwen3.6 and 64 for Gemma~4) with thinking enabled. Each cell reports mean accuracy over three runs, with the standard deviation in parentheses. BFCL uses the same decoding settings with a 32,768-token budget per call; we report its Non-Live and Live function-calling aggregates. Bold indicates the highest accuracy within each model, pruning criterion, and benchmark; underlining indicates the highest accuracy across all pruned configurations for each model and benchmark.}
\setlength{\tabcolsep}{3pt}
\begin{tabular}{llcccccc}
\toprule
 & & \multicolumn{2}{c}{Math} & \multicolumn{2}{c}{Coding} & \multicolumn{2}{c}{Tool-Use (BFCLv4)} \\
\cmidrule(lr){3-4} \cmidrule(lr){5-6} \cmidrule(lr){7-8}
Model & Method & GSM8K & MATH-500 & HE+ & MBPP+ & Non-Live & Live \\
\midrule
\multirow{10}{*}{\shortstack[c]{Qwen3.6\\(35B-A3B)}}
 & Baseline                     & 0.962 \dk{0.006} & 0.983 \dk{0.003} & 0.917 \dk{0.006} & 0.969 \dk{0.003} & 0.884 \dk{0.003} & 0.811 \dk{0.008} \\
\cmidrule(l){2-8}
 & Mass                         & 0.925 \dk{0.009} & 0.966 \dk{0.006} & 0.888 \dk{0.010} & 0.825 \dk{0.022} & 0.793 \dk{0.011} & 0.680 \dk{0.004} \\
 & \quad SlimWise               & \textbf{0.946} \dk{0.003} & \underline{\textbf{0.971}} \dk{0.003} & 0.907 \dk{0.008} & 0.883 \dk{0.020} & 0.803 \dk{0.003} & 0.740 \dk{0.001} \\
 & \quad + Distill.             & \textbf{0.946} \dk{0.007} & 0.967 \dk{0.003} & \textbf{0.921} \dk{0.005} & \textbf{0.919} \dk{0.012} & \textbf{0.872} \dk{0.005} & \textbf{0.792} \dk{0.007} \\
\cmidrule(l){2-8}
 & EAN                          & 0.941 \dk{0.008} & 0.961 \dk{0.003} & 0.900 \dk{0.012} & 0.930 \dk{0.018} & 0.829 \dk{0.004} & 0.730 \dk{0.004} \\
 & \quad SlimWise               & 0.949 \dk{0.003} & 0.959 \dk{0.003} & \textbf{0.921} \dk{0.005} & \textbf{0.957} \dk{0.001} & 0.842 \dk{0.006} & 0.766 \dk{0.007} \\
 & \quad + Distill.             & \textbf{0.960} \dk{0.003} & \textbf{0.967} \dk{0.010} & 0.917 \dk{0.015} & 0.936 \dk{0.020} & \underline{\textbf{0.877}} \dk{0.003} & \underline{\textbf{0.804}} \dk{0.003} \\
\cmidrule(l){2-8}
 & REAP                         & 0.949 \dk{0.006} & 0.964 \dk{0.008} & 0.902 \dk{0.010} & 0.932 \dk{0.003} & 0.857 \dk{0.003} & 0.759 \dk{0.002} \\
 & \quad SlimWise               & 0.959 \dk{0.004} & \underline{\textbf{0.971}} \dk{0.006} & \underline{\textbf{0.929}} \dk{0.008} & \underline{\textbf{0.967}} \dk{0.003} & 0.859 \dk{0.010} & 0.784 \dk{0.003} \\
 & \quad + Distill.             & \underline{\textbf{0.962}} \dk{0.002} & 0.969 \dk{0.001} & 0.909 \dk{0.005} & 0.946 \dk{0.003} & \underline{\textbf{0.877}} \dk{0.003} & \textbf{0.802} \dk{0.005} \\
\midrule
\multirow{10}{*}{\shortstack[c]{Gemma~4\\(26B-A4B)}}
 & Baseline                     & 0.971 \dk{0.004} & 0.962 \dk{0.002} & 0.949 \dk{0.006} & 0.979 \dk{0.004} & 0.831 \dk{0.003} & 0.807 \dk{0.002} \\
\cmidrule(l){2-8}
 & Mass                         & 0.914 \dk{0.004} & 0.896 \dk{0.003} & 0.917 \dk{0.008} & 0.945 \dk{0.001} & 0.824 \dk{0.003} & 0.690 \dk{0.006} \\
 & \quad SlimWise               & 0.938 \dk{0.003} & 0.906 \dk{0.004} & 0.907 \dk{0.012} & 0.954 \dk{0.007} & \underline{\textbf{0.825}} \dk{0.004} & 0.722 \dk{0.007} \\
 & \quad + Distill.             & \textbf{0.955} \dk{0.003} & \underline{\textbf{0.945}} \dk{0.005} & \underline{\textbf{0.947}} \dk{0.006} & \underline{\textbf{0.967}} \dk{0.003} & 0.805 \dk{0.007} & \textbf{0.773} \dk{0.005} \\
\cmidrule(l){2-8}
 & EAN                          & 0.816 \dk{0.007} & 0.549 \dk{0.012} & 0.246 \dk{0.044} & 0.343 \dk{0.031} & 0.007 \dk{0.003} & 0.012 \dk{0.005} \\
 & \quad SlimWise               & 0.173 \dk{0.044} & 0.445 \dk{0.043} & 0.494 \dk{0.054} & 0.608 \dk{0.006} & 0.052 \dk{0.006} & 0.063 \dk{0.009} \\
 & \quad + Distill.             & \underline{\textbf{0.965}} \dk{0.001} & \textbf{0.843} \dk{0.007} & \textbf{0.917} \dk{0.010} & \textbf{0.929} \dk{0.008} & \textbf{0.734} \dk{0.001} & \textbf{0.721} \dk{0.003} \\
\cmidrule(l){2-8}
 & REAP                         & 0.744 \dk{0.019} & 0.763 \dk{0.028} & 0.762 \dk{0.020} & 0.804 \dk{0.021} & 0.388 \dk{0.011} & 0.258 \dk{0.006} \\
 & \quad SlimWise               & 0.845 \dk{0.025} & 0.736 \dk{0.011} & 0.858 \dk{0.013} & 0.841 \dk{0.027} & 0.662 \dk{0.003} & 0.663 \dk{0.010} \\
 & \quad + Distill.             & \textbf{0.963} \dk{0.004} & \textbf{0.921} \dk{0.005} & \textbf{0.919} \dk{0.006} & \textbf{0.958} \dk{0.006} & \textbf{0.809} \dk{0.009} & \underline{\textbf{0.797}} \dk{0.001} \\
\bottomrule
\end{tabular}
\label{tab:sampled-std}
\end{table}

\end{document}